\pdfoutput=1

\documentclass[11pt]{article}

\usepackage[]{ACL2023}

\usepackage{times}
\usepackage{latexsym}

\usepackage[T1]{fontenc}

\usepackage[utf8]{inputenc}

\usepackage{microtype}

\usepackage{inconsolata}
\usepackage{graphicx}
\usepackage{todonotes}

\usepackage{booktabs}
\usepackage{xspace}
\newcommand{\repo}{\url{https://anonymous.4open.science/r/cost-continuous-training-EC4B}}
\newcommand{\emission}{kgCO$_2$eq\xspace}
\newcommand{\subtt}[1]{\textsubscript{\textsc{#1}}}
\usepackage[compact]{titlesec}

\title{Is It Worth the (Environmental) Cost? \\ Limited Evidence for the Benefits of Diachronic Continuous Training}
\title{Is It Worth the (Environmental) Cost? \\ Limited Evidence for Temporal Adaptation via Continuous Training}

\author{Giuseppe Attanasio$^{\clubsuit}$, Debora Nozza$^{\clubsuit}$, Federico Bianchi$^{\diamondsuit}$, Dirk Hovy$^{\clubsuit}$ \\ \\
 $^{\clubsuit}$Bocconi University, Milan, Italy \\
 $^{\diamondsuit}$Stanford University, Stanford, CA, USA \\
 { \tt \{giuseppe.attanasio3,debora.nozza,dirk.hovy\}@unibocconi.it}  \\
 \texttt{fede@stanford.edu}}

\begin{document}
    \maketitle
\begin{abstract}

Language is constantly changing and evolving, leaving language models to become quickly outdated.
Consequently, we should continuously update our models with new data to expose them to new events and facts. However, that requires additional computing, which means new carbon emissions.
Do any measurable benefits justify this cost?
This paper looks for empirical evidence to support continuous training. We reproduce existing benchmarks and extend them to include additional time periods, models, and tasks. 
Our results show that the downstream task performance of temporally adapted English models for social media data do \textit{not} improve over time. Pretrained models \textit{without} temporal adaptation are actually significantly more effective and efficient. However, we also note a lack of suitable temporal benchmarks.
Our findings invite a critical reflection on when and how to temporally adapt language models, accounting for sustainability.\footnote{Source code available at \repo.}
\end{abstract}

\section{Introduction}

Language models (LMs) are trained on a static data sample fixed in time. Because BERT \citep{devlin-etal-2019-bert} was trained on the 2019 version of Wikipedia, it does not know that Argentina won FIFA's 2022 World Cup or that the SARS-CoV-2 coronavirus has caused a pandemic.
It is intuitively problematic if language technologies do not acknowledge recent facts; old knowledge about the world might lead to uninformed outputs, or even downright false claims. Thus, recent research proposed a temporal adaptation strategy: continuous training with upstream language modeling on updated data \citep{loureiro-etal-2022-timelms,lazaridou2021mind,rottger-pierrehumbert-2021-temporal-adaptation}. We refer to the resulting models as Diachronically Adapted Language Models (DALMs).

One method, proposed by \citet{loureiro-etal-2022-timelms}, is to regularly (re-)train on recent data to keep the models up to date. While the authors argue that this strategy guarantees language models’ capacity to better deal with future out-of-distribution data, continuous training requires continuous (and escalating) computation and incurs clear environmental costs \cite{strubell-etal-2019-energy}. Adapting a pretrained RoBERTa model \citep{liu2019roberta} with 4M instances four times a year produces $\sim25$ kg CO$_2$, equivalent to driving an internal combustion engine car for $\sim230$ km (see \S\ref{ssec:tweeteval_emissions}). More recent LMs have orders of magnitude more parameters, and potentially exponentially higher emissions.
Given these non-negligible costs, we ask, ``\textit{Is there sufficient evidence to support continuous training and its environmental cost?}''




\begin{figure}[!t]
\centering
\includegraphics[width=\linewidth]{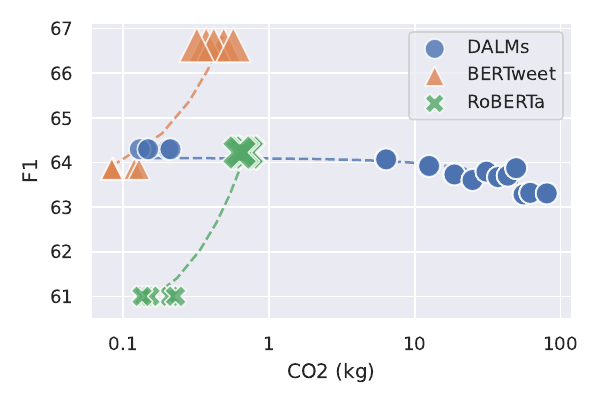}
\caption{Carbon footprint vs.\ downstream performance on \textsc{TweetEval} benchmark \citep{barbieri-etal-2020-tweeteval}. Marker size correspond to model size (\textsc{base}, \textsc{large}).}
\label{fig:emission-vs-f1}
\end{figure}

We address this question by focusing on the continuous training strategy proposed by \newcite{loureiro-etal-2022-timelms}, who update their models with the latest Twitter data every three months.
We reproduce and extend their original setup by 
(1) testing the downstream performance of \textit{all} DALMs released to date to study performance across time periods; 
(2) assessing whether performance is impacted if training and evaluation data are temporally aligned; and 
(3) comparing DALMs to pretrained LMs without temporal adaptation, inspecting the tradeoff between performance, model size, and emissions.


Our empirical results show that 1) downstream task performance does \textit{not} improve significantly over time and 2) temporal alignment between models and data does \textit{not} provide any improvement.
Moreover, simple pretrained LMs match or outperform DALMs at a fraction of the environmental cost. Figure~\ref{fig:emission-vs-f1} reports the carbon footprint vs.\ performance of RoBERTa~\cite{liu2019roberta}, BERTweet \citep{nguyen-etal-2020-bertweet}, and DALMs on the main task in \citet{loureiro-etal-2022-timelms}.

\paragraph{Contributions} We reproduce the experiments of \citet{loureiro-etal-2022-timelms} to extensively evaluate DALMs from all time frames, finding no significant improvement over time at high environmental costs.
Moreover, we demonstrate that pretrained LMs without temporal adaptation are more effective and efficient than DALMs, and show no benefits when models and data are temporally aligned.

\section{Diachronic Continuous Training}
\label{sec:dct}

Diachronic Continuous Training (DCT) is a form of temporal adaptation of LMs \citep{loureiro-etal-2022-timelms,rottger-pierrehumbert-2021-temporal-adaptation}. DCT uses an upstream objective, e.g., Masked Language Modeling, and data from more recent periods than the original training data.

\citet{loureiro-etal-2022-timelms} proposed to run DCT recurrently and incrementally: Every period $t$, DALM$_{t}$ is created by running MLM on DALM$_{t-1}$ with new data from period $[t-1,t]$.
Specifically, they created a first checkpoint by adapting a pretrained RoBERTa model \citep{liu2019roberta} on a dataset of 90M English tweets, sampled through 2019 (\textsc{DALM 19}). Next, they trained and released quarterly checkpoints running DCT on 4.2M additional tweets sampled each quarter. Here, we denote them with \textsc{YY-Q} (e.g., 20-3).\footnote{Please refer to the original paper for the full details on the data collection and training procedures.}


While DCT is a straightforward adaptation strategy, it requires additional computation and entails environmental costs \citep{strubell-etal-2019-energy}.
To justify this cost, we investigate three research questions (RQs).
(RQ1) Does DALMs' downstream performance improve over time? Ideally, learning from new events would create \textit{better} models, especially if temporally aligned with the test data. \\
(RQ2) Do DALMs outperform pretrained LMs? If that is not the case, we argue DALMs might not have a proper use case.\\
(RQ3) Are DALMs (environmentally) sustainable? I.e., considering how performance evolves over time and compared to existing models, are they worth the cost? 






\begin{table*}[!th]
\centering
\begin{tabular}{@{}lrrrrrrrr@{}}
\toprule
\textbf{Model} & \textbf{Emoji} & \textbf{Emotion} & \textbf{Hate} & \textbf{Irony} & \textbf{Offensive} & \textbf{Sentiment} & \textbf{Stance} & \textbf{Avg} \\ \midrule



RoBERTa\subtt{base} & 30.76 & 78.81 & 39.53 & 69.84 & 78.66 & 71.14 & 58.28 & 61.00 \\
RoBERTa\subtt{large} & 33.78 & 81.64 & 47.04 & 75.43 & 79.68 & 70.90 & 61.10 & 64.22 \\
BERTweet\subtt{base} & 31.30 & 80.92 & 50.00 & 79.45 & 80.38 & 71.85 & 53.12 & 63.86 \\
BERTweet\subtt{large} & \textbf{34.83} & \textbf{82.46} & 49.44 & \textbf{83.46} & 78.66 & 71.96 & \textbf{63.62} & \textbf{66.64} \\
DALM 19 & 31.98 & 81.55 & \textbf{50.55} & 74.75 & 81.36 & 71.81 & 58.10 & 64.30 \\ \midrule
20-1 & 0.50 & -0.24 & -1.15 & -1.91 & \textbf{0.30} & 0.01 & 0.90 & -0.23 \\
20-2 & 0.58 & -0.36 & -1.62 & -1.13 & -0.01 & 0.12 & -0.20 & -0.38 \\
20-3 & 0.54 & -0.63 & -2.18 & -1.49 & -0.04 & 0.24 & -0.40 & -0.57 \\
20-4 & 0.50 & -0.39 & -2.14 & -1.79 & -0.94 & 0.22 & -0.30 & -0.69 \\
21-1 & 0.65 & -0.56 & -2.68 & -1.34 & -1.08 & 0.21 & 1.28 & -0.50 \\
21-2 & 0.59 & -0.37 & -4.07 & -1.18 & -0.70 & 0.07 & 1.28 & -0.62 \\
21-3 & 0.68 & 0.21 & -3.54 & -1.48 & -0.97 & 0.00 & 0.97 & -0.59 \\
21-4 & 0.60 & -0.51 & -3.75 & -0.69 & -0.57 & 0.04 & 1.91 & -0.42 \\
22-1 & 0.44 & -0.15 & -5.18 & -0.99 & -0.82 & 0.15 & -0.53 & -1.01 \\
22-2 & 0.59 & -0.58 & -4.80 & -0.72 & -1.01 & 0.24 & -0.57 & -0.98 \\
22-3 & 0.52 & -1.16 & -4.67 & -0.43 & -0.24 & \textbf{0.31} & -1.26 & -0.99 \\ \bottomrule

\end{tabular}
\caption{Mean F1 macro ($\uparrow$) on \textsc{TweetEval} by classification task on 5 initialization seeds. Results of diachronic models (20-1 through 22-3) are relative to the first checkpoint (19-4). Best model per task in bold. Standard deviation and significance details in Appendix.}
\label{tab:tweeteval}
\end{table*}

\section{Experiments}

We reproduce the setup in \citet{loureiro-etal-2022-timelms} and extend the evaluation to \textit{all} DALMs the authors released to date. This set includes intermediate and more recent checkpoints until September 2022, for a total of twelve models.

We compare DALMs to existing pretrained RoBERTa \citep{liu2019roberta} and BERTweet \citep{nguyen-etal-2020-bertweet}, a variant adapted to Twitter, in two size configurations, \textsc{base} and \textsc{large}. We do not temporally adapt either of them.

We report training details and estimated emissions of our experiments in Appendix \ref{app:sec:training} and \ref{app:sec:footprint}.

\subsection{RQ1. Does DALMs' downstream performance improve over time?}
\label{ssec:tweeteval}

We replicate the same experimental setup used by  \newcite{loureiro-etal-2022-timelms}, testing the models on the \textsc{TweetEval} benchmark \citep{barbieri-etal-2020-tweeteval}. The suite collects seven existing English Twitter classification datasets.\footnote{There are both binary and multi-class tasks, please refer to \citet{barbieri-etal-2020-tweeteval} for further details.}
\newcite{loureiro-etal-2022-timelms} evaluate the performance using only DALM 19 and 21-4, showing that the latter is competitive even on past tweets and that time-aware training is relevant.

Table~\ref{tab:tweeteval} reports the results for \textit{all} released DALMs.\footnote{We average macro F1 of the five stance detection-related tasks.} Top rows represent baseline models without temporal adaptations.
The bottom rows correspond to the remaining DALMs. 
Crucially, \textbf{DALMs' downstream task performance does not improve over time}. In five tasks (Emotion, Hate, Irony, Offensive, and Stance), DALM 19 outperforms 22-3; whereas in Emoji and Sentiment, 22-3's improvements over DALM 19 are lower than 2\%. This finding contrasts with the intended use of DCT---improving language models over time---and with results in \citet{loureiro-etal-2022-timelms}. 

Moreover, DALMs are unstable and generally lead to worse performance when adapted over time. DALM 22-1, adapted with 15 months' worth of new data, performs worst, with a 1.6\% F1 decrease, while 22-3 loses 1.5\% F1 on average compared to DALM 19.

\begin{figure}[!t]
\centering
\includegraphics[width=.9\linewidth]{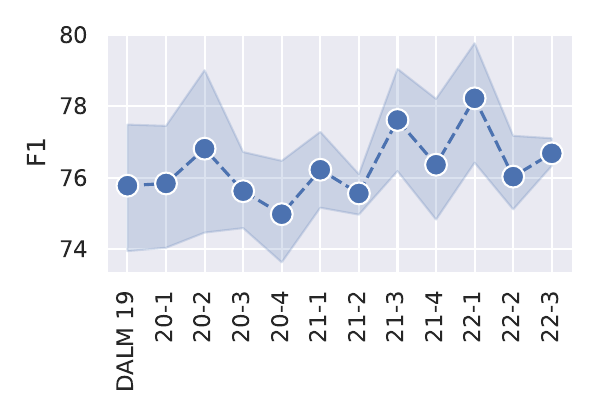}
\caption{DALMs' F1 (macro) performance on \textsc{Grimm}. 95\% confidence intervals over 5 initialization seeds.}
\label{fig:grimminger}
\end{figure}

\subsubsection{Testing on Temporally Aligned Data}

Although the \textsc{TweetEval} task was used to demonstrate DALM capabilities in \citet{loureiro-etal-2022-timelms}, all tasks had data prior to the training period of the first DALM, i.e., 2019. Here, we investigate whether DALMs are useful when the model and testing data are concurrent (i.e., temporally aligned).
We test DALMs on the Twitter Hate Speech dataset \textsc{Grimm} \citep{grimminger-klinger-2021-hate}. The dataset contains 3,000 tweets about the 2020 US election from six weeks before to one week after election day (Nov. 20, 2020). Tweets are annotated for (binary) hatefulness/offensiveness detection. We report additional details in Appendix~\ref{sec:datasets-details}.


Figure~\ref{fig:grimminger} shows the results. Surprisingly, DALM 20-4 performs worst, despite being most temporally aligned with the evaluation data. We hypothesize that tweets from 2020's last quarter (immediately after the election) act as a confounding factor, rather than additional meaningful knowledge.
This result matches prior work reporting \textbf{no benefits from a temporal alignment between models and testing data} \citep{rottger-pierrehumbert-2021-temporal-adaptation}, further contradicting the intended use of DALMs.

 

\subsection{RQ2. Do DALMs outscore existing pretrained LMs?}


The primary objective for DALMs is to improve on existing models that are static in time.
However, our empirical experiments on \textsc{TweetEval} (Table \ref{tab:tweeteval}) reveal that this is not the case.
BERTweet\subtt{large} \textit{without temporal adaptation} is the best model on average, matching (Hate, Offensive, and Sentiment) or outperforming (Emoji, Emotion, Irony, and Stance) DALMs.


This finding suggests that \textbf{it is more beneficial to train a single model with more capacity and generalization capabilities once than to adapt smaller models over time.}

We note how domain adaptation on Twitter data leads to higher performance: BERTweet and DALMs are significantly better than RoBERTa\subtt{base}, although marginally better than RoBERTa\subtt{large}. 

\subsubsection{RQ3. Are DALMs sustainable?}
\label{ssec:tweeteval_emissions}

Despite optimized hardware efficiency, training large-scale models incurs substantial energy consumption \citep{strubell-etal-2019-energy,10.1145/3442188.3445922}.
We extend our analysis by relating downstream task performance and carbon emissions.





We compute carbon footprint as follows.
For DALMs, we consider DALM 19 as a ``pre-existing'' reference. Since each DALM is built upon the preceding one, we consider the emissions produced by the \textit{n}-th model to be \textit{n} times the emission of a single DCT run, plus its fine-tuning cost.
For pretrained models (RoBERTa and BERTweet), we consider the CO$_2$ emissions from fine-tuning only.
Figure~\ref{fig:emission-vs-f1} shows the relationship between carbon footprint and downstream task. See Appendix~\ref{sec:dalm-emissions} for extended results and details.

BERTweet\subtt{large} performs best, producing 0.43 kgCO$_2$. DALM 20-1 produces 15 times that amount, while scoring 2.57 F1 points less on average. DCT also shows diminishing returns: Training DALM 22-3 requires 80.38 kgCO$_2$ (187 times that of BERTweet\subtt{large}), without significant improvement.
For reference, 80.38 \emission is equivalent to driving an internal combustion engine car for 745 km.\footnote{Estimate based on average CO$_2$ emissions of new passenger cars in EU in 2020 from \url{https://www.acea.auto/figure/average-co2-emissions-of-new-cars-in-eu/}}
These results highlight that pretrained LMs are drastically more efficient and that DALMs are provably less sustainable than a ``one model, multiple use cases'' approach.



\section{Related Work}

Recent work has addressed the issue of large language models becoming outdated, focusing on benchmarking their use upstream and downstream over time, and proposing solutions to update them.

\citet{lazaridou2021mind} question generalization on future data, showing that Transformer-XL models underperform in predicting utterances from beyond their training period, and that increasing size does not solve the issue.
\citet{jang2022temporalwiki} studies how factual knowledge evolves in LMs, showing that perplexity improves over subsequent Wikipedia dumps.
\citet{rottger-pierrehumbert-2021-temporal-adaptation} compare temporal to domain adaptation with Reddit data. While temporal adaptation is generally beneficial upstream, the authors found limited evidence of its benefits over domain adaptation on downstream tasks.

Research on evaluation proposes benchmarks to test pretrained models over new events and facts \citep{dhingra-etal-2022-time,jang2022temporalwiki}, and social media data \citep{rottger-pierrehumbert-2021-temporal-adaptation,loureiro-etal-2022-tempowic}. 
\citet{dhingra-etal-2022-time} propose a temporal benchmark to evaluate factual knowledge in T5 \citep{JMLR:v21:20-074} models adapted yearly on news data. 

Parallel research investigates temporal adaptation strategies. Some model time explicitly: \citet{dhingra-etal-2022-time} prepend time information as strings for sequence-to-sequence span completion, \citet{10.1145/3488560.3498529} prepend a time token to input sequences, \citet{rosin-radinsky-2022-temporal} add a new time component in the Transformer attention.
\citet{de-cao-etal-2021-editing} learn an external network to modify the model's weights at inference time and improve LMs in knowledge-intensive mask-filling and question-answering tasks.


\section{Conclusion}

We compare the downstream performance of twelve Diachronically Adapted Language Models \cite[DALMs,][]{loureiro-etal-2022-timelms} over time to two pretrained models. Our experiments show that DALMs' performance does \textit{not} improve over time, nor do they become better than pretrained models.


All our findings point in the same direction: There is very limited evidence to justify the costs of temporal adaptation via continuous training. We provide two possible interpretations. The easiest one is that continuous training does not impact the model's ability to transfer knowledge to downstream tasks. In that case, we should prioritize different ways to account for time in LMs \citep[e.g.,][]{dhingra-etal-2022-time,de-cao-etal-2021-editing}.
Alternatively, currently used benchmarks to validate DALMs do not have a real temporal use case compared to temporally-aligned setups \citep{lazaridou2021mind,rottger-pierrehumbert-2021-temporal-adaptation}. In that case, failure or success in solving these tasks says little about continuous training as a method for temporal adaptation.

Our results invite a profound reflection on whether the proven environmental cost justifies an unproven assumption. The solution might require better time-aligned evaluation paradigms, adaptation strategies, or both.

\section*{Limitations}


We have replicated the evaluation setup for downstream performance from \citep{loureiro-etal-2022-timelms}. However, this benchmark might not provide full insights into the performance of temporally adapted models. For example, none of the tasks has data temporally aligned with DALMs.
Although our analysis on the more recent \textsc{Grimm} \citep{grimminger-klinger-2021-hate} dataset showed no correlation between training and test set alignment, we cannot exclude that insights might change using different datasets or downstream tasks. 

Further, our setup is influenced by the focus on Twitter data, which might not be representative of \textit{all} world events, the passing of time, or the introduction of novel words. We believe Twitter, and social media in general, is a good proxy for these aspects, but we might find different results in different domain setups.


\bibliography{anthology,custom}
\bibliographystyle{acl_natbib}

\appendix





\section{Details on Experiments}
\label{app:sec:details-experiments}

\subsection{Training Setup}
\label{app:sec:training}

We fine-tune and test our models on the \textsc{TweetEval} (Section~\ref{ssec:tweeteval}) separately per task, using the official splits. For \textsc{Grimm}, we use the official training and test splits and hold out 10\% of training data as validation, stratifying on the target class distribution. 

For a fair comparison, we use the same training budget and hyperparameter setup for all the models.
We use pretrained models and tokenizers from the HuggingFace Hub \citep{wolf-etal-2020-transformers}. RoBERTa and BERTweet \textsc{base} and \textsc{large} variants count 110M and 335M learnable parameters, respectively.
Hyper-parameters are reported in Table~\ref{tab:hparams}.
We use the checkpoint with the best validation loss for testing. Note that best models were always achieved before 10 epochs. We repeat experiments on 5 different initialization seeds. 

\begin{table}[!t]
\centering
\begin{tabular}{@{}ll@{}}
\toprule
\textbf{Hyper-parameter} & \textbf{Value} \\ \midrule
Maximum sequence length & 256 \\
Batch size equivalent & 32 \\
Max training epochs & 10 \\
Peak learning rate & 2e-5 \\
Learning rate scheduler & Linear decay \\
Warmup steps & 10\% \\
Weight decay & 0.01 \\
Float precision & fp16 \\ 
Evaluation steps & 500 (20 for Stance \\ & Detection tasks) \\
Monitored metric & Validation loss \\ \bottomrule
\end{tabular}
\caption{Hyper-parameter values for training tasks.}
\label{tab:hparams}
\end{table}


\begin{table*}[!ht]
\centering
\begin{tabular}{@{}lrrrrrrrr@{}}
\toprule
\textbf{Model} & \textbf{Emoji} & \textbf{Emotion} & \textbf{Hate} & \textbf{Irony} & \textbf{Offensive} & \textbf{Sentiment} & \textbf{Stance} & \textbf{Avg} \\ \midrule
RoBERTa\subtt{base} & 0.50 & 0.45 & \textbf{1.87} & 3.65 & 1.31 & 0.60 & 2.94 & 1.62 \\
RoBERTa\subtt{large} & 1.42 & 0.49 & 4.27 & 2.97 & 1.83 & 0.89 & 3.68 & 2.22 \\ 
BERTweet\subtt{base} & \textbf{0.31} & 0.34 & 2.79 & 2.26 & 0.88 & 0.98 & 3.62 & 1.60 \\
BERTweet\subtt{large} & 1.33 & \textbf{0.27} & 4.44 & 5.57 & 0.95 & 0.59 & 4.56 & 2.53 \\
DALM 19 & 1.33 & 0.51 & 3.01 & \textbf{0.77} & 1.05 & 0.46 & 3.80 & 1.56 \\
20-1 & 0.55 & 0.70 & 2.00 & 1.87 & 1.81 & 0.61 & 3.72 & 1.61 \\
20-2 & 0.53 & 0.56 & 3.11 & 1.71 & 1.51 & 0.34 & 4.87 & 1.80 \\
20-3 & 0.60 & 0.51 & 3.16 & 1.43 & 1.90 & 0.29 & 4.18 & 1.73 \\
20-4 & 0.53 & 0.49 & 3.00 & 1.27 & 1.64 & 0.38 & 3.64 & 1.56 \\
21-1 & 0.48 & 0.61 & 2.15 & 1.52 & 1.97 & 0.45 & 3.06 & 1.46 \\
21-2 & 0.49 & 0.90 & 3.08 & 1.17 & 2.46 & 0.34 & 2.93 & 1.62 \\
21-3 & 0.44 & 1.01 & 2.39 & 1.74 & 1.46 & 0.43 & 2.68 & 1.45 \\
21-4 & 0.56 & 0.64 & 3.35 & 1.51 & 1.48 & 0.22 & \textbf{2.36} & \textbf{1.45} \\
22-1 & 0.76 & 0.51 & 3.13 & 1.75 & 1.87 & 0.24 & 4.23 & 1.79 \\
22-2 & 0.70 & 1.12 & 2.77 & 1.87 & 1.31 & \textbf{0.19} & 3.95 & 1.70 \\
22-3 & 0.58 & 0.74 & 3.65 & 2.07 & \textbf{0.71} & 0.44 & 3.88 & 1.72 \\ \bottomrule
\end{tabular}
\caption{F1 (macro) standard deviation ($\downarrow$) on \textsc{TweetEval} by classification task on 5 initialization seeds. Best model per task in bold.}
\label{tab:tweeteval-std}
\end{table*}

\subsection{Statistical Significance}

\begin{table}[!ht]
\centering
\begin{tabular}{@{}lrrr@{}}
\toprule
\textbf{Task} & \multicolumn{1}{l}{\textbf{DALM 19}} & \multicolumn{1}{l}{\textbf{21-4}} & \multicolumn{1}{l}{\textbf{22-3}} \\ \midrule
Emoji & $\bullet$ & $\bullet$ & $\bullet$ \\
Emotion & - & - & $\bullet$ \\
Hate & - & - & $\bullet$ \\
Irony & $\bullet$ & $\bullet$ & $\bullet$ \\
Offensive & - & - & - \\
Sentiment & - & - & - \\
Stance & $\bullet$ & $\bullet$ & $\bullet$ \\ \bottomrule
\end{tabular}
\caption{Statistical significance of BERTweet\subtt{large}'s F1 (macro) improvement over DALM 19, 21-4, and 22-3. $\bullet$: p < 0.01}
\label{tab:significance}
\end{table}

We test the statistical significance via bootstrap sampling \citep{sogaard-etal-2014-whats} using the open-source \texttt{boostsa} python library.\footnote{\url{https://github.com/fornaciari/boostsa}}
We compute significance on every task of the \textsc{TweetEval} benchmark using 1000 bootstrap samples and a sample size of 20\% of the test set. 

We report in Table~\ref{tab:significance} the statistical significance of performance between BERTweet\subtt{large} (the best-performing model on average) and DALM 19 (first checkpoint released in \citet{loureiro-etal-2022-timelms}, 21-4 (last), and 22-3 (last model released to date). Compared to DALM 19, we record a significant improvement only on Stance for 21-4 and no improvements at all for 22-3.


\subsection{Carbon Footprint}
\label{app:sec:footprint}

Our experiments were conducted using private hardware of type RTX A6000 PCIe 24GB (TDP 230W).
Total emissions are estimated to be 17.18 kgCO$_2$eq.
    
Estimations were conducted using the CodeCarbon library \cite{lacoste2019quantifying}.\footnote{\url{https://github.com/mlco2/codecarbon}}

\section{Dataset Details}
\label{sec:datasets-details}

Both \textsc{TweetEval} and \textsc{Grimm} are released publicly free of restrictions. We respect their intended use---i.e., training and testing automatic detection models of specific attributes from Twitter data.

\subsection{\textsc{Grimm}}

\citet{grimminger-klinger-2021-hate} collected 382,210 tweets about the 2020 US presidential election using keywords targeting the two candidates (e.g., \#Trump2020, \#TrumpPence2020, \#Biden2020, \#BidenHarris2020). Tweets were sampled within the period from six weeks before to one week after the election day (November 20, 2020).

The authors sampled 3,000 and annotated them for hateful or offensive language and political stances. In this paper, we focus on the former. The proposed Hate Speech Detection task is a text classification task with binary labels. Class distribution is highly imbalanced (Hateful: 88\%).
We used the official training and test sets for our tests. 

\section{Additional Results}

\subsection{\textsc{TweetEval}}

We thoroughly followed the training details in \citet{loureiro-etal-2022-timelms} to reproduce results on DALMs. The authors' \textit{TimeLM 19} and \textit{TimeLM 21} correspond to our DALM 19 and 21-4, respectively.
\citeauthor{loureiro-etal-2022-timelms} do not state whether they average results across different seeds. We run all models with 5 different random weight initializations for a more robust evaluation. Table~\ref{tab:tweeteval-std} reports the F1 (macro) standard deviation across 5 different initialization seeds.

Our results do not match exactly those reported by the authors but are within two (RoBERTa\subtt{base}, DALM 19, and 21-4) or three (BERTweet\subtt{base}) standard deviations on average.
We attribute this result to variance in model initialization. For example, note that our DALM 19 gains $\sim27$ F1 points compared to \textit{TimeLM 19} on the task of Irony detection.

We highlight that larger models (RoBERTa\subtt{large}, BERTweet\subtt{large}) also have a larger standard deviation in results. We attribute this result to hyperparameter selection and specifically to the fact that some parameters---coming from the original 
 setup in \citet{loureiro-etal-2022-timelms}---might be optimized for smaller models.

\subsection{DALMs' Carbon Emission}
\label{sec:dalm-emissions}

\begin{table}[!th]
\centering
\begin{tabular}{@{}lrr@{}}
\toprule
\textbf{Model} & \textbf{kgCO$_2$} & \textbf{F1} \\ \midrule
RoBERTa\subtt{base} & 0.17 & 61.00 \\
RoBERTa\subtt{large} & 0.69 & 64.22 \\
BERTweet\subtt{base} & 0.10 & 63.86 \\
BERTweet\subtt{large} & 0.43 & 66.64 \\
DALM 19 & 0.17 & 64.30 \\
20-1 & 6.34 & 64.07 \\
20-2 & 12.51 & 63.92 \\
20-3 & 18.68 & 63.73 \\
20-4 & 24.85 & 63.61 \\
21-1 & 31.02 & 63.80 \\
21-2 & 37.19 & 63.68 \\
21-3 & 43.36 & 63.71 \\
21-4 & 49.53 & 63.88 \\
22-1 & 55.70 & 63.29 \\
22-2 & 61.87 & 63.32 \\
22-3 & 80.38 * & 63.31 \\ \bottomrule
\end{tabular}
\caption{Mean carbon emission (continuous training plus fine-tuning) and F1 (macro) across 5 initialization seeds on \textsc{TweetEval}. *: \href{https://huggingface.co/cardiffnlp/twitter-roberta-base-sep2022}{22-3} has been trained with a 15M tweet increment.}
\label{tab:dalms-emissions}
\end{table}

Table~\ref{tab:dalms-emissions} reports carbon emissions and average classification performance on \textsc{TweetEval} for all the models studied.

We estimate each continuous training run in \citet{loureiro-etal-2022-timelms} has produced 6.17 kgCO$_2$. We derive this estimate as follows. Based on the indication provided by the authors, we estimate each DCT round on 4.2M tweets to take 134.4 compute hours on one NVIDIA V100 PCIe 16GB (TDP 300W). We assume training is carried out on private infrastructure in the United Kingdom and thus set a carbon intensity of 153 gCO$_2$eq from \url{https://app.electricitymaps.com/zone/GB}. Estimates do not consider any CO$_2$ emission offsets.

\end{document}